\documentclass{article}

\usepackage{arxiv}

\usepackage[utf8]{inputenc}
\usepackage[T1]{fontenc}
\usepackage{hyperref}
\usepackage{url}
\usepackage{booktabs}
\usepackage{amsfonts}
\usepackage{nicefrac}
\usepackage{microtype}
\usepackage{graphicx}
\usepackage{amsmath}
\usepackage{amssymb}
\usepackage{multirow}
\usepackage{adjustbox}

\graphicspath{ {./images/} }

\title{ZipMVS: Multi-View Stereo with Compressed Cost Volumes}
\author{
  Guanglin Jin \\
  School of Artificial Intelligence and Robotics, Hunan University, Changsha 410082, China \\
  \texttt{jglinn@hnu.edu.cn} \\
  \And
  Hongshan Yu\textsuperscript{*} \\
  School of Artificial Intelligence and Robotics, Hunan University, Changsha 410082, China \\
  \texttt{yuhongshancn@hotmail.com} \\
  \And
  Javier Civera \\
  I3A, Universidad de Zaragoza, Zaragoza 50018, Spain \\
  \texttt{jcivera@unizar.es} \\
  \And
  Zhaoxin Li \\
  Agricultural Information Institute, Chinese Academy of Agricultural Sciences, Beijing 100081, China \\
  Key Laboratory of Agricultural Big Data, Ministry of Agriculture and Rural Affairs, Beijing 100081, China \\
  \texttt{cszli@hotmail.com}
}

\begin{document}

\maketitle

\begin{abstract}
Multi-view stereo (MVS) methods typically deliver highly accurate 3D reconstructions from multiple registered RGB images, thanks to the highly informative, geometric constraints between them. However, their substantial memory requirements remain a major obstacle for deployment in domains such as aerospace and autonomous systems, where resource efficiency is critical.
In this work, we introduce ZipMVS, an MVS method specifically designed for efficient high-quality reconstruction. We propose a novel depth-hypothesis strategy that enables substantial compression of the cost volume, hence greatly reducing GPU memory consumption while preserving reconstruction accuracy.
Experiments on the DTU and Tanks and Temples datasets show that ZipMVS achieves competitive reconstruction quality compared with other efficiency-oriented MVS methods, while achieving a competitive balance between reconstruction quality and GPU memory usage. 
The code is available at \url{https://github.com/JihnGlyn/ZipMVS}.
\end{abstract}

\section{INTRODUCTION}

Multi-View Stereo (MVS) is a fundamental task in 3D computer vision with broad applicability across numerous real-world scenarios. 
Given a set of RGB images captured from different viewpoints, together with their associated camera poses and intrinsics, the goal of MVS is to recover a dense representation of the scene’s 3D geometry, typically in the form of a point cloud~\cite{seitz2006comparison,furukawa2015multi,wang2024learning}. 

One of the main barriers to deploying MVS algorithms in autonomous unmanned systems is their computational footprint. Traditional handcrafted pipelines~\cite{furukawa2015multi}, represented by the PatchMatch method, offer lightweight operation suitable for resource-constrained platforms, but their reliance on local features makes them sensitive to challenging lighting or low-texture conditions. Learning-based MVS approaches~\cite{wang2024learning}, while more computationally demanding, have become the dominant choice due to their robustness and accuracy. In particular, deep learning–based methods consistently achieve strong performance even in complex environments~\cite{mvsnet, p-mvsnet, point-mvsnet}.

Modern MVS pipelines typically rely on 3D convolutions over a plane-sweep cost volume \(V\in \mathbb{R}^{C\times D\times H\times W}\) to establish reliable multi-view correspondences. However, the memory requirements of these cost volumes are substantial, often limiting their deployment in embedded or real-time systems. In this work, we directly address this bottleneck by compressing the cost volume through a reduction in the number of depth hypotheses. Although fewer hypotheses may slightly degrade reconstruction accuracy due to a narrower exploration of the depth range, the resulting memory savings are crucial for many practical applications.

Most MVS methods that aim to reduce GPU memory rely either on iterative refinement~\cite{wang2022itermvs, patchmatchnet} or binary-search-style depth sampling~\cite{gbi}. This raises a natural question: can we maintain high reconstruction quality with a principled depth-hypothesis strategy that avoids iteration and repeated searches at each level, using only a small number of depth samples? 
We argue that blindly increasing the capacity of deep networks is not a meaningful path toward more efficient MVS. Inspired by~\cite{cascade}, we instead propose a more deliberate depth-sampling strategy combined with lightweight networks, to achieve high-quality reconstructions with substantially reduced memory consumption.

In traditional coarse-to-fine pipelines, the coarse-stage depth map is upsampled and used to initialize subsequent stages, where progressively narrower sampling intervals refine accuracy. This approach, however, struggles to correct erroneous coarse-stage predictions. PatchmatchNet~\cite{patchmatchnet} partially alleviates this limitation by leveraging the spatial continuity of depth across neighboring pixels, combining priors from both coarse predictions and local propagation to recover from early mistakes.

We conceptualize depth hypothesis generation as governed by two complementary philosophies: a conservative strategy that incrementally refines predictions from the coarsest level, and an exploratory strategy that proposes depth alternatives suggested by neighboring pixels to escape from potential local minima at the coarsest level. By synergistically combining these two paradigms, our framework avoids exhaustive depth sweeping, hence improving computational efficiency, while maintaining reconstruction fidelity.

Our main contributions are two-fold. We propose a pixel-wise adaptive depth sampling strategy with a differentiable scaling network that adjusts depth ranges. We also introduce a GRU-based Depth Speculator that aggregates spatial cues to generate reliable depth hypotheses, reducing sample count without sacrificing accuracy. Additionally, a synthetic space dataset is constructed as an extra validation scenario for low-texture environments.

\section{RELATED WORK}
\subsection{Handcrafted Methods}
Classical MVS approaches can be broadly categorized into four families: patch-based~\cite{patch1, patch2}, surface-based~\cite{surface1, surface2, surface3}, voxel-based~\cite{voxel1, voxel2}, and depth map-based~\cite{gipuma, sfm, ACMM} methods. While the first three aim to directly recover full 3D models, their complex optimization pipelines and heavy computational requirements often hinder their practical deployment in real-world environments.

Depth map-based approaches strike a more favorable balance between accuracy and efficiency. These methods estimate a depth map for each reference view independently and subsequently fuse the maps into a unified point cloud.
Among depth map-based methods, Gipuma~\cite{gipuma} and COLMAP~\cite{sfm} pioneered parallel propagation and joint optimization of depth and normals. Subsequent works refine hypothesis evaluation via adaptive sampling~\cite{ACMM, hpm-mvs}, planar priors~\cite{Xu2020ACMP, tapa}, deformable patches~\cite{APD-MVS}, or semantic segmentation~\cite{msp-mvs}, achieving robustness in textureless or large-scale scenes.

A parallel line of work focuses on improving the quality of hypothesis evaluation, which is crucial in ambiguous or textureless regions. ACMP~\cite{Xu2020ACMP} and TapaMVS~\cite{tapa} incorporate planar priors to better constrain hypothesis scoring.
APD-MVS~\cite{APD-MVS} introduces deformable evaluation regions that selectively cover reliable pixels, offering strong robustness and memory efficiency.
MSP-MVS~\cite{msp-mvs} leverages Semantic-SAM for multi-granularity depth-edge aggregation and introduces adaptive sector division with disassemble-clustering to refine patch deformation and further boost reconstruction quality.

\subsection{Learning-based Methods}
Modern learning-based MVS methods typically construct cost volumes from multi-view warped features and regularize them using 3D convolutions to regress per-pixel depth maps~\cite{izquierdo2025mvsanywhere}. However, the significant computational and memory overhead of 3D convolutions often requires the use of reduced-resolution cost volumes.

Several strategies have been proposed to mitigate these limitations. R-MVSNet~\cite{R-MVSNet} processes sequential 2D cost slices with GRUs to reduce memory consumption, albeit at the cost of increased runtime. CasMVSNet~\cite{cascade} introduces a coarse-to-fine cascade of cost volumes built from feature pyramids, while UCS-Net~\cite{ucs-net} employs uncertainty-guided variance estimates to construct thin, adaptive volumes.
CVP-MVSNet~\cite{cvp-mvsnet} jointly builds pyramids of both images and cost volumes, yielding improved multi-scale representations. PatchmatchNet~\cite{patchmatchnet} embeds the PatchMatch algorithm directly into a hierarchical structure, enabling fast and memory-efficient depth estimation.
IterMVS~\cite{wang2022itermvs} models per-pixel depth distributions using a lightweight probability estimator, achieving improved performance with reduced computation.
CDS-MVSNet~\cite{cds-mvsnet} explicitly handles matching ambiguities and visibility constraints, while its half-resolution processing strategy drastically reduces resource usage without compromising accuracy. LLR-MVSNet~\cite{llr-mvsnet} replaces standard 2D convolutions with depthwise separable convolutions, cutting the number of parameters significantly while preserving high-quality reconstruction.

These approaches can be broadly grouped into four categories: (1) simplifying network architectures by removing heavy 3D convolutions or RNNs~\cite{patchmatchnet,mvs2d}, (2) reducing the spatial dimensions \textit{H} and \textit{W} of cost volumes through coarse-to-fine frameworks~\cite{cascade, cvp-mvsnet}, (3) reducing the depth dimension \textit{D} by lowering the number of depth samples~\cite{ucs-net, gbi}, and (4) reducing the channel dimension \textit{C} by grouping features~\cite{gwc}. Our method jointly adopts the latter three strategies to further alleviate GPU memory consumption.

\section{METHOD}

\begin{figure}[htbp]
    \centering
    \includegraphics[width=\textwidth]{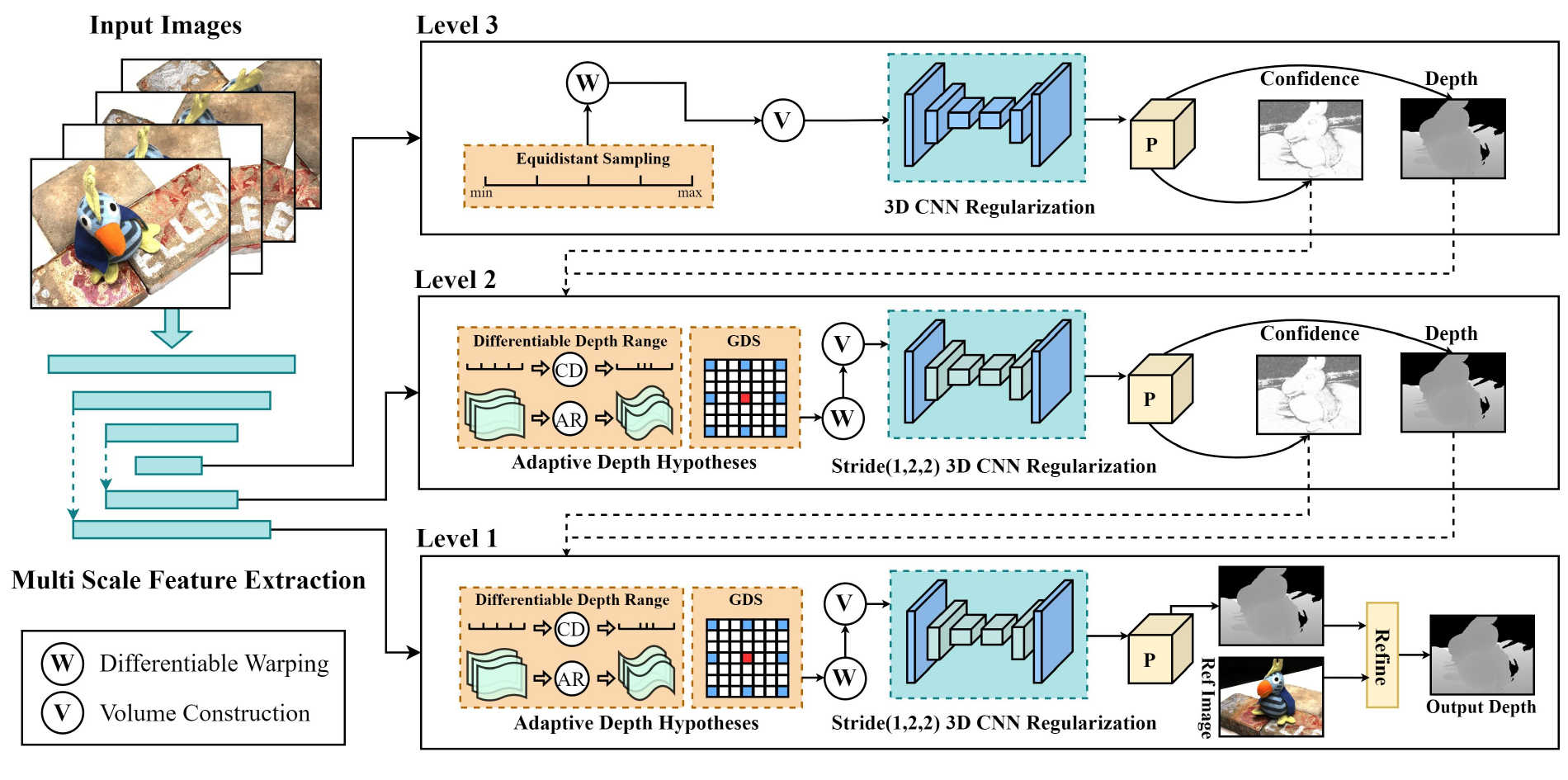}
    \caption{Overview of ZipMVS. Multi-view images are first processed by the feature extractor. Depth estimation begins at level 3, and in subsequent levels, new depth hypotheses are generated using the confidence and depth maps from the previous level. The final refined depth map is produced as the output.}
    \label{fig:network}
\end{figure}

Figure~\ref{fig:network} shows an overview of the architecture of ZipMVS. Its three main processing modules start with the extraction of multi-scale features. These are used for the hierarchical depth hypotheses with which the cost volume is built. The next subsections detail each of these modules, together with the loss that we minimize to train them. 
Differentiable warping refers to the standard operation of projecting source features onto reference view under each depth hypothesis, and volume construction denotes the group-wise correlation and weighted aggregation. These operations follow prior work~\cite{gwc, pvsnet} and are not novel contributions of this paper.

\subsection{Multi-scale Feature Extractor}
The input consists of a reference image \(I_0\in \mathbb{R}^{H\times W\times3}\) and \(N-1\) source images \(\{I_i\}_{i=1}^{N-1}\), each with camera parameters \(c_0,...,c_N\). Our Feature Pyramid Network (FPN)~\cite{fpn} outputs multi-scale features \(\textbf{F}_{i,l}\) for image \(i\) at level \(l\), with spatial resolution \(1/2^l\) and channels \(C=16,32,64\) for \(l=1,2,3\).

\subsection{Adaptive Depth Hypotheses}
At the coarsest level (\(l=3\)), the full depth range is uniformly discretized. At finer levels (\(l=2,1\)), we use the Differentiable Depth Range (DDR) and GRU-based Depth Speculator (GDS) modules.

\subsubsection{Differentiable Depth Range}

\begin{figure}[htbp]
    \centering
    \includegraphics[width=\textwidth]{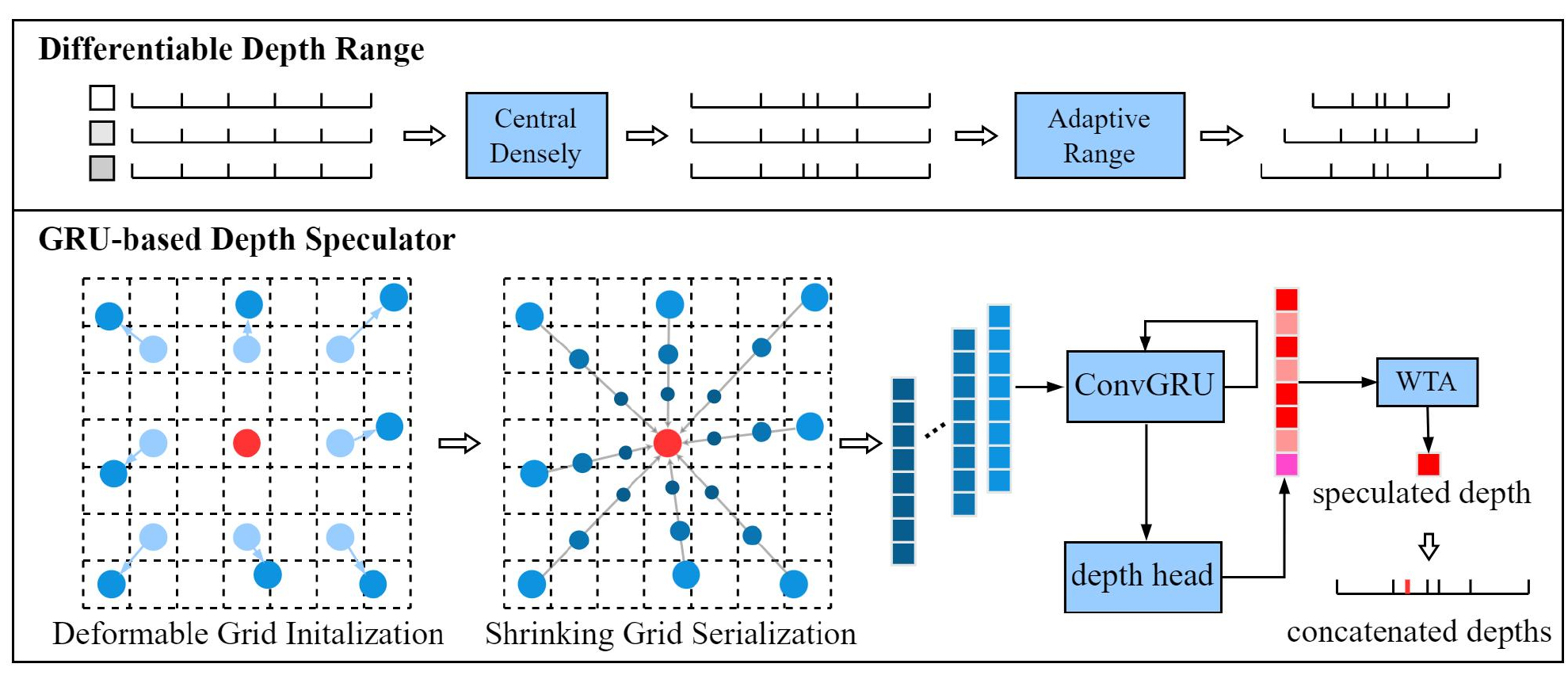}
    \caption{Demonstration of the Advanced Depth Hypothesis Strategy, which contains a Differentiable Depth Range module (DDR) and a GRU-based Depth Speculator module (GDS).}
    \label{fig:ADH}
\end{figure}

Regions with high photometric confidence need narrower depth intervals, while low-confidence regions benefit from wider intervals. Thus, we propose the DDR module with an Adaptive Range Network (ARNet) that predicts a pixel-wise depth-range map from the previous level's confidence. We also introduce a Central Densely (CD) sampling strategy: depth planes are concentrated near the interval center and become sparser toward the boundaries, keeping symmetry.

Our formulation is intentionally simple. We take the central value of the sampling interval to be the depth estimate from the previous level. The distance of each depth plane from the center increases quadratically, producing symmetric yet progressively sparser depth samples toward both ends of the interval. The quadratic coefficient is predicted by ARNet using the confidence map as input. ARNet consists of three Conv–BN–ReLU blocks followed by a final Conv2D layer without activation, with channel dimensions $1 \rightarrow 8 \rightarrow 16 \rightarrow 8 \rightarrow 1$. All convolutional layers use a kernel size of 3. To prevent the sampling range from becoming excessively large or vanishingly small, the output is clamped to \([\zeta_{min},\zeta_{max}]\). We set $\zeta_{\min}=0.5$ and $\zeta_{\max}=5.0$ based on empirical observation of the depth range variations in the DTU dataset. These values ensure the sampling interval is neither too narrow to miss the true depth nor too wide to waste hypotheses.
We adopt symmetric sampling because the true depth can lie on either side of the prior estimate with equal likelihood. A fixed step size would bias the search, whereas quadratic spacing focuses more samples near the center (where the prior is most confident) while still covering a wide range with fewer samples.

Let $\theta_l$ denote the base depth interval scale at level $l\in\{1,2\}$, which is a learnable scalar shared across all pixels. Let $A_l \in \mathbb{R}^{H_l \times W_l \times 1}$ be the quadratic coefficient map predicted by ARNet from the confidence map of the previous level. For each pixel $(u,v)$, the depth sampling interval is centered at the upsampled depth from level $l+1$, denoted as $D_{l+1}(u,v)$. We sample $n$ hypotheses in inverse-depth space to maintain linearity along the viewing ray. Let $x_s$ be linearly spaced values from $-1$ to $1$ with $n$ points: $x_s = \{-1, -1+\frac{2}{n-1}, \dots, 1\}$. The signed quadratic offset $o_s$ is computed element-wise as:
\begin{align}
o_s &= \text{sign}(x_s) \cdot x_s^2, \\
\delta_s &= A_l(u,v) \cdot o_s,\\
d_l^{(s)}(u,v) &= \frac{1}{U(D_{l+1})(u,v)} + \theta_l \cdot \left(\frac{1}{d_{\min}} - \frac{1}{d_{\max}}\right) \cdot \delta_s,
\end{align}
where $d_{\min}, d_{\max}$ are the global depth bounds, $U(\cdot)$ denotes bilinear upsampling, and the final depth is obtained by $d = 1/\text{(inverse depth)}$. All operations are pixel-wise and broadcasted.

\subsubsection{GRU-based Depth Speculator}
We propose a GRU (Gated Recurrent Unit)-based Depth Speculator (GDS). Inspired by PatchmatchNet~\cite{patchmatchnet}, we also assume that the depth of neighboring pixels provide useful priors. However, directly propagating these values to construct the cost volume often limits the quality of the reconstruction. Instead, analogous to curvature estimation~\cite{cds-mvsnet}, our method enables each pixel to perceive local depth variations in a simpler and more structured manner. Instead of directly relying on raw neighbor-propagated depths, we progressively infer the central depth from its surroundings using a sequence prediction network that extracts depth cues from spatial neighborhoods and produces higher-quality depth hypotheses.
Specifically, each pixel samples uniformly spaced depths along 4 or 8 directions; the outermost value initializes a GRU, which iteratively predicts inward until reaching the center. Offsets of the outermost samples are learned via a deformable convolution.

For each pixel, we sample depths along 4 or 8 directions (determined by the level) from its spatial neighbors. In each direction, we take 4 points at equal intervals from the pixel to the boundary. The outermost point's offset is learned via a deformable convolution. Let $x_k$ be the feature vector at step $k$ (from outermost to center), which is the concatenation of the pixel's own feature and the sampled depth value from the neighbor at that step. We use a convolutional GRU that processes the sequence $x_0, x_1, ..., x_n$ in order. The hidden state $h_k$ is updated by:
\begin{align}
z_k &= \sigma(\text{Conv}([h_{k-1}, x_k], W_z)), \\
r_k &= \sigma(\text{Conv}([h_{k-1}, x_k], W_r)), \\
\hat{h}_k &= \tanh(\text{Conv}([r_k \odot h_{k-1}, x_k], W_h)), \\
h_k &= (1 - z_k) \odot h_{k-1} + z_k \odot \hat{h}_k. \\
P &= \text{Softmax}(\text{Conv}(h_k)). \notag
\end{align}
After the final step $k=3$, a depth head (three 2D convolutions) maps $h_3$ to a speculative depth value for each pixel. For each of the 4 (or 8) sampled directions, we obtain a candidate speculative depth. We then from a probability volume and apply winner-take-all strategy across these directional candidates—selecting the one with the highest likelihood as the final speculative depth for the pixel. The final depth hypothesis set is the union of DDR-generated depths and the speculative depth, resulting in $n+1$ candidates per pixel.

\subsection{Cost Volume Construction}

We warp source-view feature maps into the reference view under multiple depth hypotheses, enabling the construction of multi-scale cost volumes.

Let \textbf{p} denote a pixel in the reference image \(I_0\), and let \(\textbf{p}_{i,j}\) represent the projection of \(\textbf{p}\) onto the source view \(I_i\) under the \(j\)-th depth hypothesis, i.e., \(d_{i,j}\).
Given known intrinsics \(\textbf{K}_i\), relative rotation \(\textbf{R}_{0\rightarrow i}\) and translation \(\textbf{t}_{0\rightarrow i}\) between the reference view and source views, \(\textbf{p}_{i,j}\) is computed as:
\begin{equation}
    \textbf{p}_{i,j}= d_{i,j}^{-1}\cdot\textbf{K}_i\cdot(\textbf{R}_{0\rightarrow i }\textbf{K}_0^{-1} \textbf{p}\cdot d_{0,j} + \textbf{t}_{0\rightarrow i}).
\end{equation}

For each view \(i\), warped feature representations are obtained via bilinear interpolation according to the \(j\)-th depth hypothesis and denoted as \(\textbf{F}_i(\textbf{p}_{i,j})\), while the reference feature maps are denoted as \(\textbf{F}_{0}(\textbf{p})\).

The feature channels \(C\) are uniformly divided into \(G\) groups, and group-wise correlation~\cite{gwc} is applied to compute the cost for each depth hypothesis. Multi-view aggregation is then performed using learned pixel-wise view weights~\cite{sfm, pvsnet}.
Denoting \(\textbf{F}_{i}^{g}\) as the \(g\)-th feature group of \(\textbf{F}_{i}\), the \(i\)-th cost volume \(\textbf{V}_i\) is computed as:
\begin{equation}
    \textbf{V}_{i}^{g}(\textbf{p},j)=\frac{G}{C}\langle \textbf{F}_{0}^{g}(\textbf{p}),  \textbf{F}_{i}^{g}(\textbf{p}_{i,j})\rangle.
\end{equation}
where \(\langle \cdot, \cdot \rangle\) denotes the dot product operation. To eliminate the group dimension, we take the average across all $G$ groups for each pixel and depth hypothesis. This group-wise correlation reduces computational overhead and enables more efficient handling of the cost volume.

View weights are predicted using three consecutive \(1\times1\times1\) 3D convolution layers, followed by a sigmoid activation. A maximum operation along the depth dimension \(D\) produces the final view weight \(\textbf{w}_{i}(\textbf{p})\) for pixel \(\textbf{p}\) in source image \(I_i\). This pixel-wise view weight is reused in subsequent levels via bilinear upsampling.
The final group-wise matching cost \(\Bar{\textbf{V}}(\textbf{p},j)\) is computed as a weighted sum:
\begin{equation}
    \Bar{\textbf{V}}(\textbf{p},j) = \frac{\Sigma _{i=1}^{N-1}\textbf{w}_{i}(\textbf{p}) \cdot \textbf{V}_{i}(\textbf{p},j)}{\Sigma _{i=1}^{N-1}\textbf{w}_{i}(\textbf{p})}.
\end{equation}

The weighted feature volume \(\Bar{\textbf{V}}(\textbf{p},j)\) is processed by a 3D-UNet cost volume regularization network to generate a probability volume \(\textbf{P}(\textbf{p},j)\). At level 3, we use \(3\times3\times3\) kernels for the 3D-UNet cost volume regularization network. In the subsequent levels, given that the depth dimension of the cost volume is small (with only 5 depth samples), the downsampling stride is set to \((1,2,2)\), preserving depth information. 

Given that depth values are sampled at unequal intervals, the confidence map is calculated by summing the probabilities of the three closest depth hypotheses per pixel. The regressed depth \(\textbf{D}(\textbf{p})\) is then computed as:
\begin{equation}
    \textbf{D}(\textbf{p})=\sum_{j=1}^{D}d_j\cdot \textbf{P}(\textbf{p},j).
\end{equation}

The final-level depth map has resolution \(\frac{H}{2}\times \frac{W}{2}\). We refine it using the full-resolution RGB reference image \(H\times W\) via a residual refinement network, identical to the one in PatchmatchNet~\cite{patchmatchnet}, which predicts a residual to correct the upsampled depth map.

\section{Experiments}
\subsection{Datasets}
We evaluate our approach in three public datasets typically used in the multi-view stereo literature, plus an additional in-house synthetic dataset simulating the conditions of space applications.
\subsubsection{Public Benchmarks}
The DTU dataset~\cite{dtu} contains multiple views of indoor scenes with corresponding camera poses, comprising 124 unique environments captured under seven lighting conditions. Following SurfaceNet~\cite{surfacenet}, we adopt the standard training, validation, and testing splits.
We also use Tanks and Temples~\cite{tnt}, a large-scale benchmark that includes intermediate and advanced subsets. Performance on this dataset is evaluated by submitting reconstructed point clouds to the official online server.

\subsubsection{In-House Synthetic Data}
To evaluate our approach in a space environment, we generated a synthetic scenario in Blender\footnote{https://blender.org}. We imported NASA's 3D model of the International Space Station\footnote{https://nasa3d.arc.nasa.gov/models} and placed it at the coordinate origin.
Image acquisition and camera parameter extraction were automated using the software by Cartucho \emph{et al.}~\cite{visionblender}.
We used a standard pinhole camera model with a focal length of 53.70mm and an output image resolution of \(900\times1200\) pixels.
The sequence taken by the camera was uniformly sampled along the trajectory to capture 20 images. Rendering was performed with the Cycles engine using a noise threshold of $0.01$ and a maximum of $1024$ samples.
Note that this dataset is intended as a proof-of-concept for space applications, where the camera motion is assumed to be accurately known (e.g., from spacecraft navigation systems). The uniform black background simulates the emptiness of space, which, while extreme, tests the method's ability to handle low-texture scenes. We do not claim this dataset as a major contribution, but rather as an additional validation scenario.

\subsection{Implementation Details}
\subsubsection{Training Details}
We train our model on the DTU training set, using input images resized to \(512\times 640\) and \(N=5\) views. Features are partitioned into groups of \(8,8,8\) across three pyramid levels. The number of depth hypotheses for levels \(l=1,2,3\) is set to \(4,4,48\). At level \(l=3\), the full depth range is uniformly discretized, while for levels \(l=2\) and \(l=1\). The starting radii and the number of sampling neighbors in the GDS module are configured as \(6, 4\) and \(8, 4\) respectively. 

We optimize the entire network using the Adam optimizer in PyTorch for 24 epochs. The initial learning rate is $0.001$ and is reduced by a factor of 1.5 after epochs \(12,14,16,18,20\) and \(22\). All training experiments use a batch size of 4 on a single NVIDIA RTX 4090 GPU. Additionally, we use a robust selection strategy for the source images mentioned in \cite{patchmatchnet} that randomly choose four from the ten best source views.

\subsubsection{Evaluation Details}
All evaluation settings are consistent with those used at training. As our model adopts a three-level architecture and the cost-volume regularization module is also implemented as a three-level U-Net, with each level performing downsampling, the feature maps are reduced to half their size at every level. To prevent size mismatches during upsampling, the input resolution must therefore be divisible by \(2^{6}\). The DTU test images have a resolution of \(1200\times 1600\), which is not divisible by 64. To address this, we introduce a simple size-adjustment step (nearest-neighbor interpolation) within the cost-volume regularization network, rather than resizing the original images to \(1152\times 1600\). Depth estimation is performed on images at their native resolution of \(1200\times 1600\). Outliers are filtered using both photometric and geometric consistency, and all depth maps of a scene are subsequently fused into a single point cloud. 

\subsection{Experimental Results}
\subsubsection{Results on DTU}

On DTU, we estimate depth maps using \(N=4\) input views at \(1200\times1600\) resolution. At levels $2$ and $1$, the initial depth interval ratios are set to 0.05 and 0.02 respectively, the initial radii of the GDS module are still set as 6 and 4 respectively.
We conduct quantitative assessment by computing accuracy and completeness, employing official MATLAB codes. As shown in Table \ref{tab:1}, compared to other MVS methods that do not adjust hyperparameters according to specific scenarios, our method stands out with its competitive performance regarding overall error performance. 
Our accuracy (Acc.) of 0.369mm is slightly worse than GBi-Net (0.315mm) but better than IterMVS (0.373mm), indicating competitive precision. Our completeness (Comp.) of 0.284mm is competitive, though third after GBi-Net (0.262mm) and PatchmatchNet (0.277mm). The overall error of 0.327mm ranks second among all methods, demonstrating that ZipMVS achieves a favorable balance between accuracy and completeness.
\begin{table}[htbp]
    \caption{Results for baselines and ZipMVS on DTU. The lower the better for accuracy (Acc.), completeness (Comp.), and overall. Best result is bolded, and second-best is in italic bold. The first horizontal line separates traditional (above) from learning-based (below) methods.}
    \label{tab:1}
    \centering
    \begin{tabular}{cccc}
    \hline
        Methods & Acc.(mm)↓ & Comp.(mm)↓ & Overall (mm)↓ \\
    \hline
        Camp~\cite{campbell2008using}& 0.935 & 0.554 & 0.695\\
        Furu~\cite{patch1} & 0.613 & 0.941 & 0.777\\
        Tola~\cite{tola2012efficient} & 0.342 & 1.190 & 0.766\\
        Gipuma~\cite{gipuma} & \textbf{0.283} & 0.873 & 0.578\\
    \hline
        MVSNet~\cite{mvsnet} & 0.396 & 0.527 & 0.462\\
        R-MVSNet~\cite{R-MVSNet} & 0.383 & 0.452 & 0.417\\
        Point-MVSNet~\cite{point-mvsnet} & 0.342 & 0.411 & 0.376\\
        CasMVSNet~\cite{cascade} & 0.325 & 0.385 & 0.355\\
        CVP-MVSNet~\cite{cvp-mvsnet} & \textbf{\textit{0.296}} & 0.406 & 0.351\\
        Vis-MVSNet~\cite{vis-mvsnet} & 0.369 & 0.361 & 0.365\\
        UCS-Net~\cite{ucs-net} & 0.338 & 0.349 & 0.344\\
        GBi-Net~\cite{gbi} & 0.315 & \textbf{0.262} & \textbf{0.289}\\
        PatchmatchNet~\cite{patchmatchnet} & 0.427 & \textbf{\textit{0.277}} & 0.352\\
        IterMVS~\cite{wang2022itermvs} & 0.373 & 0.354 & 0.363\\
    \hline
        ZipMVS (ours) & 0.369& 0.284& \textbf{\textit{0.327}}\\
    \hline
    \end{tabular}
\end{table}
\begin{figure}[htbp]
    \centering
    \includegraphics[width=0.8\linewidth]{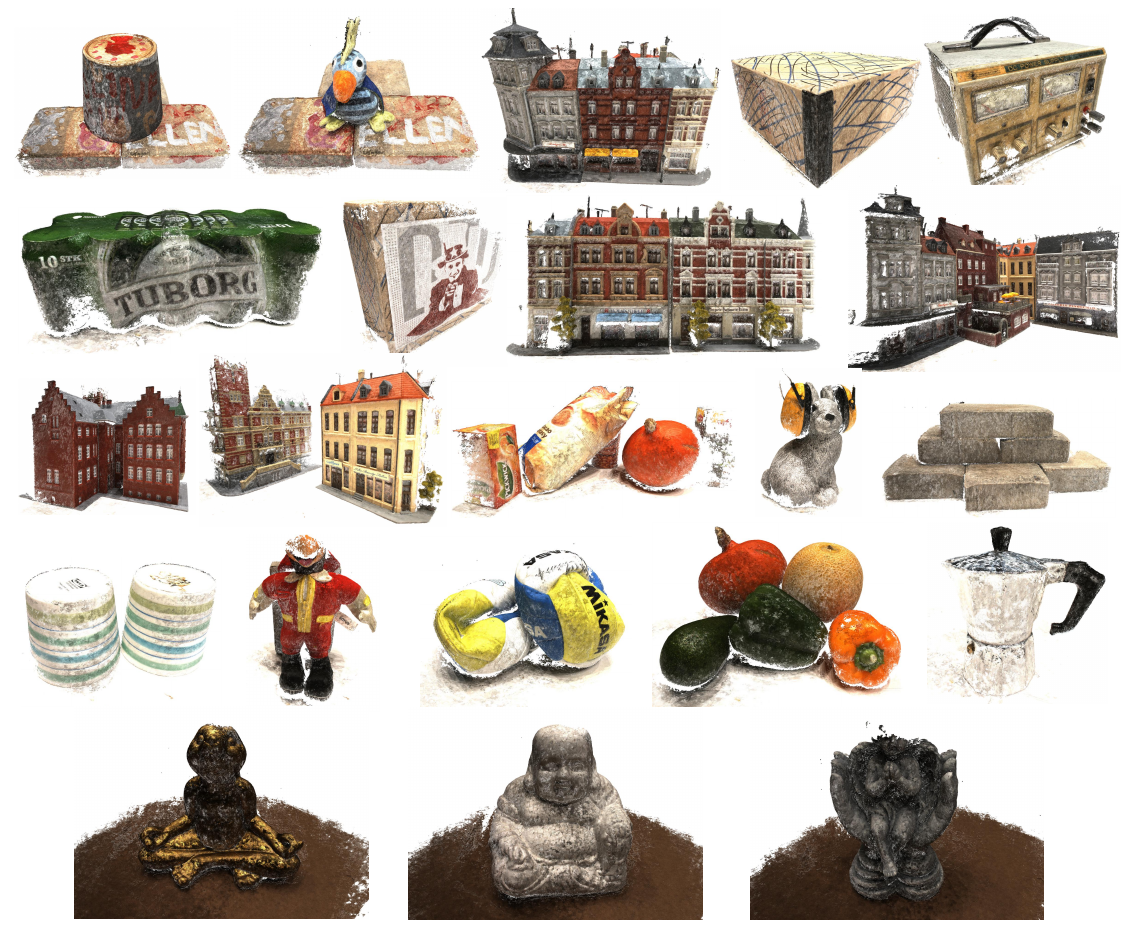}
    \caption{Qualitative reconstruction results of our ZipMVS on the 22 subsets of the DTU test set.}
    \label{fig:mem}
\end{figure}
\begin{figure}[htbp]
    \centering
    \includegraphics[width=0.8\linewidth]{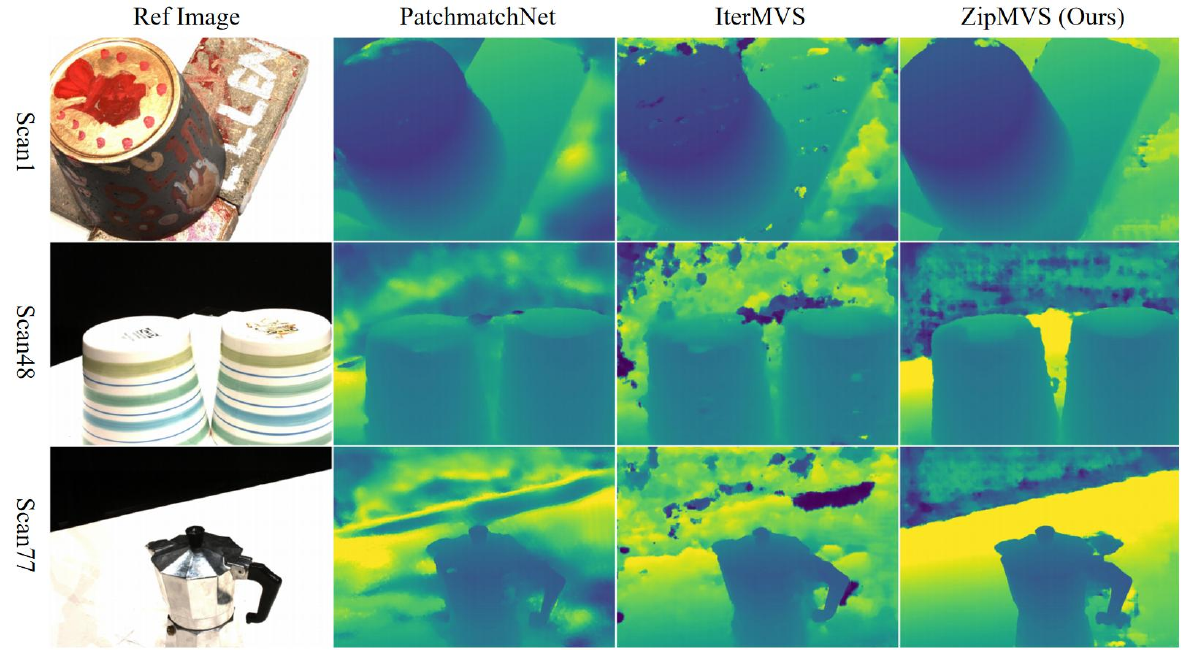}
    \caption{Depth maps generated by PatchmatchNet~\cite{patchmatchnet} ($2^\text{nd}$ column), IterMVS~\cite{wang2022itermvs} ($3^\text{rd}$ column), and our ZipMVS ($4^\text{th}$ column) for scan1 ($1^\text{st}$ row), scan48 ($2^\text{nd}$ row), and scan77 ($3^\text{rd}$ row) of the DTU dataset.}
    \label{fig:dtu_compare}
\end{figure}

\subsubsection{Results on Tanks and Temples}

The evaluation is performed on the Tanks and Temples benchmark~\cite{tnt}, with two resolution settings \(1024\times1920\) or \(1024\times2048\). 
The number of input views differs according to subsets, we set \(N=7\) for the Intermediate and \(N=11\) for the Advanced. The results for both subsets are shown in Table \ref{tab:2}. The number of depth hypotheses for level \(l=3,2,1\) are set to \([4,4,48]\), and other hyperparameters remain unchanged across scenarios.
\begin{table*}[htbp]
    \centering
    \caption{Results on Tanks and Temples for baselines trained only on DTU. The higher the better for precision, recall and F-score.}
    \label{tab:2}
    \footnotesize
    \begin{tabular}{c|ccc|ccc}
        \hline
        \multirow{2}{*}{Methods} & \multicolumn{3}{c|}{Intermediate Dataset} & \multicolumn{3}{c}{Advanced Dataset} \\
        & Precision↑ & Recall↑ & F-score↑ & Precision↑ & Recall↑ & F-score↑ \\
        \hline
        Colmap~\cite{sfm} & 43.16 & 44.48 & 42.14 & \textbf{33.65} & 23.96 & 27.24 \\
        MVSNet~\cite{mvsnet} & 40.23 & 49.70 & 43.48 & - & - & - \\
        R-MVSNet~\cite{R-MVSNet} & 43.74 & 57.60 & 48.40 & 31.47 & 22.05 & 24.91 \\
        Point-MVSNet~\cite{point-mvsnet} & 41.27 & 60.13 & 48.27 & - & - & - \\
        CasMVSNet~\cite{cascade} & 47.62 & \textbf{\textit{74.01}} & \textbf{56.84} & \textbf{\textit{29.68}} & 35.24 & 31.12 \\
        CVP-MVSNet~\cite{cvp-mvsnet} & \textbf{51.41} & 60.19 & 54.03 & - & - & - \\
        UCS-Net~\cite{ucs-net} & 46.66 & 70.34 & 54.83 & - & - & - \\
        PatchmatchNet~\cite{patchmatchnet} & 43.64 & 69.37 & 53.15 & 27.27 & 41.66 & 32.31 \\
        IterMVS~\cite{wang2022itermvs} & \textbf{\textit{46.82}} & 73.50 & \textbf{\textit{56.22}} & 28.04 & \textbf{42.60} & \textbf{33.24} \\
        ZipMVS (Ours) & 44.77& \textbf{74.79}& 54.46& 27.67& \textbf{\textit{42.38}}& \textbf{\textit{33.03}}\\
        \hline
    \end{tabular}
\end{table*}
\begin{figure}[htbp]
    \centering
    \includegraphics[width=0.8\linewidth]{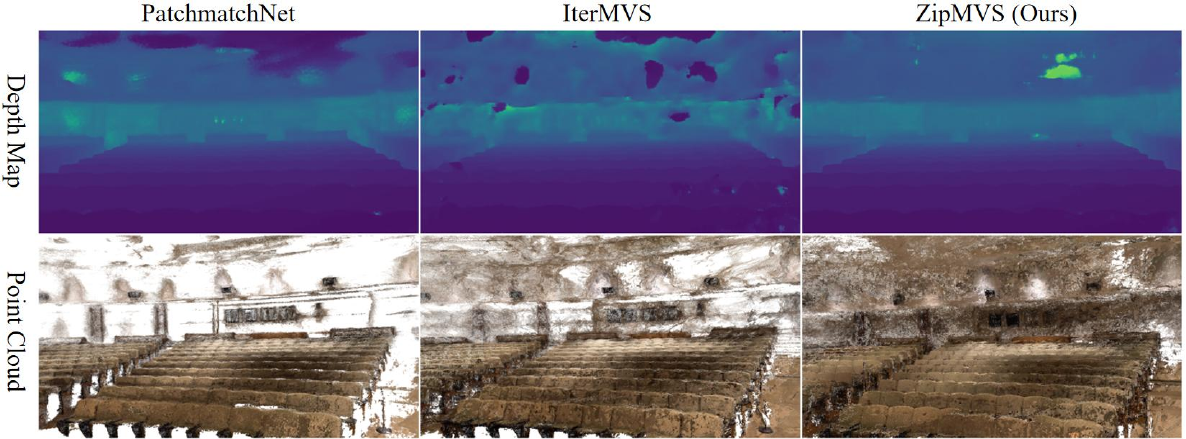}
    \caption{Depth maps (first row) and point clouds (second row) estimated by  PatchmatchNet~\cite{patchmatchnet} (left column), IterMVS~\cite{wang2022itermvs} (central column), and our ZipMVS (right column) on the Auditorium set of Tanks and Temples.}
    \label{fig:compare}
\end{figure}

On the Intermediate set, ZipMVS achieves an F-score of 54.46, slightly lower than CasMVSNet (56.84) and IterMVS (56.22). This gap is largely due to lower precision (44.77 vs. 47.62 for CasMVSNet and 46.82 for IterMVS), as our method favors completeness (74.79, the highest among all methods) over precision in large-scale scenes. On the Advanced set, our F-score of 33.03 is competitive with IterMVS (33.24) and outperforms PatchmatchNet (32.31). The high recall of 42.38 indicates that our depth hypotheses effectively cover the target geometry even in challenging outdoor scenarios.

\subsubsection{Memory Usage and Runtime}

Our evaluation compares both memory usage and run-time performance against several efficient learning-based MVS methods on the DTU testing dataset. All measurements were performed on a single NVIDIA RTX 4090 with PyTorch 1.12, using batch size 1 and input resolution 1152×1600. We use PyTorch functions to measure the peak allocated memory usage. Although we achieved the best results when N was 4, we set the number of input images N to 5 for each method for a fair comparison. We measure the run-time by taking the average of the time it takes to infer all depth maps. Due to the trade-off between iteration times and accuracy in methods such as PatchmatchNet~\cite{patchmatchnet} and IterMVS~\cite{wang2022itermvs}, we compared our method with their default evaluation parameters.

\begin{table}[htbp]
    \centering
    \caption{Comparison of several methods in terms of memory usage, run-time on the DTU dataset. 
    Bold and italic bold indicate the best and the second-best, respectively.}
    \label{tab:3}
    \footnotesize
    \begin{tabular}{c|cc|c}
    \hline
         Methods & Mem.(MB) & Runtime(s) & Overall(mm)↓\\
    \hline
         CasMVSNet~\cite{cascade} & 4591 & - & 0.355 \\
         PatchmatchNet~\cite{patchmatchnet} & 1629 & 0.139& \textbf{\textit{0.352}} \\
         IterMVS~\cite{wang2022itermvs} & \textbf{886} & \textbf{0.093}& 0.363 \\
         ZipMVS (5 views) & \textbf{\textit{1322}}& \textbf{\textit{0.112}}& \textbf{0.340}\\
    \hline
    \end{tabular} 
\end{table}

According to Table \ref{tab:3}, our model has a slightly longer runtime compared to IterMVS~\cite{wang2022itermvs}, but it is shorter than that of PatchmatchNet~\cite{patchmatchnet}. Moreover, it shows superior overall performance on the DTU dataset.
Compared to IterMVS, our method uses 49\% more memory (1322MB vs. 886MB) and runs 20\% slower (0.112s vs. 0.093s). Our efficiency gains are best appreciated relative to non-iterative baselines like MVSNet and CasMVSNet.

\begin{figure}[htbp]
    \centering
    \includegraphics[width=0.6\linewidth]{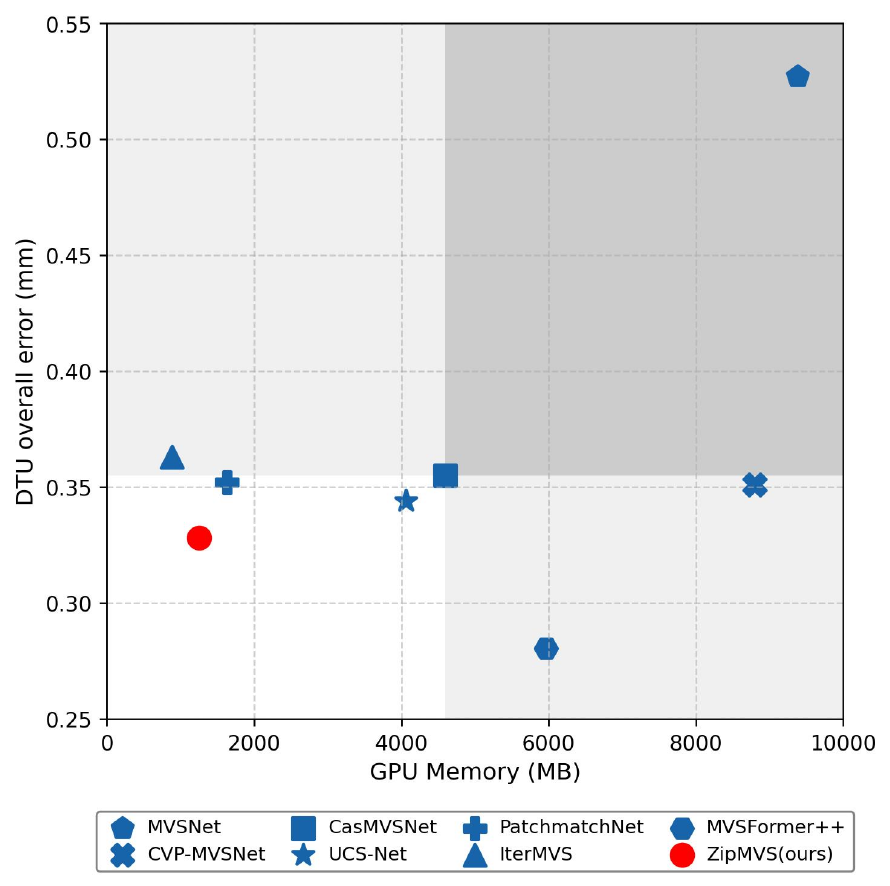}
    \caption{Overall error (mm) and GPU memory usage (MB) of several MVS baselines and our ZipMVS on the DTU dataset at an input resolution of $1152 \times 1600$. Points closer to the lower-left corner indicate a better trade-off between reconstruction quality and memory footprint. ZipMVS stands out as a practical alternative: it is surpassed only by MVSFormer++~\cite{mvsformer++}, yet requires over $3\times$ less GPU memory. Compared to the remaining baselines, ZipMVS delivers superior accuracy while achieving competitive memory efficiency compared to many existing methods.}
    \label{fig:mem}
\end{figure}

\begin{figure}[htbp]
    \centering
    \includegraphics[width=0.6\linewidth]{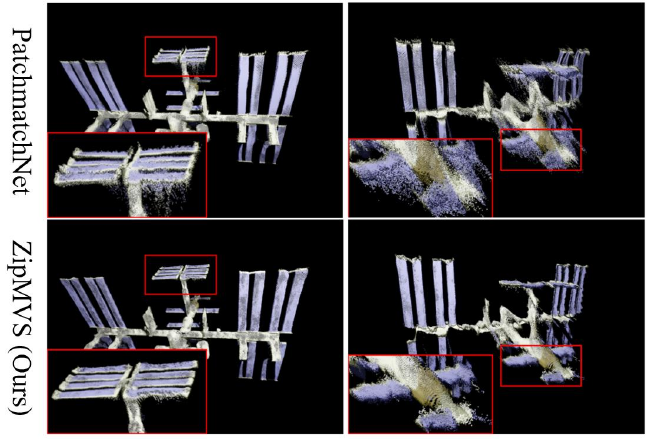}
    \caption{Reconstruction results on our in-house synthetic dataset. The first row shows the output of PatchmatchNet~\cite{patchmatchnet}, while the second row shows the reconstruction produced by our ZipMVS. Note the higher quality of ZipMVS results, in particular in the parts framed in red.}
    \label{fig:synth}
\end{figure}

\subsubsection{Evaluation on Synthetic Data}
We carried out experiments on synthesized data using both PatchmatchNet~\cite{patchmatchnet} and our proposed ZipMVS. The input images are original size \(900\times1200\), and the number of views \(N\) is set to 5. Figure~\ref{fig:synth} shows that ZipMVS achieves superior reconstruction quality compared to PatchmatchNet. 
In terms of efficiency, ZipMVS requires 824MB and 0.073s per depth map, compared to 1024MB and 0.078s for PatchmatchNet, demonstrating improved quality with lower resource consumption.

\subsection{Ablation Study}
\subsubsection{Adaptive Range \& Central Densely Interval}
We compare several ablated variants of ZipMVS with the full model on the DTU test set. Specifically, we evaluate three ablation settings: removing the AR module, removing the CD module, and removing both. All models were trained for 24 epochs using identical hyperparameters. As shown in Table~\ref{ARCD}, both modules contribute positively to the overall performance.

\begin{table}[htbp]
    \centering
    \caption{Ablation study evaluating the impact of the Adaptive Range (AR) and Central Densely Interval (CD) modules on depth hypotheses generation.}
    \begin{tabular}{c|ccc}
    \hline
         Methods &  Acc.(mm)↓&  Comp.(mm)↓& Overall(mm)↓\\
    \hline
         w/o AR \& CD&  0.383&  0.383& 0.383\\
         w/o AR &  \textbf{0.352}&  0.351& 0.352\\
         w/o CD &  0.362&  0.366& 0.364\\
          ZipMVS (ours) &  0.369& \textbf{0.284}& \textbf{0.327}\\
    \hline
    \end{tabular}
    \label{ARCD}
\end{table}

\begin{figure}[htbp]
    \centering
    \includegraphics[width=0.6\linewidth]{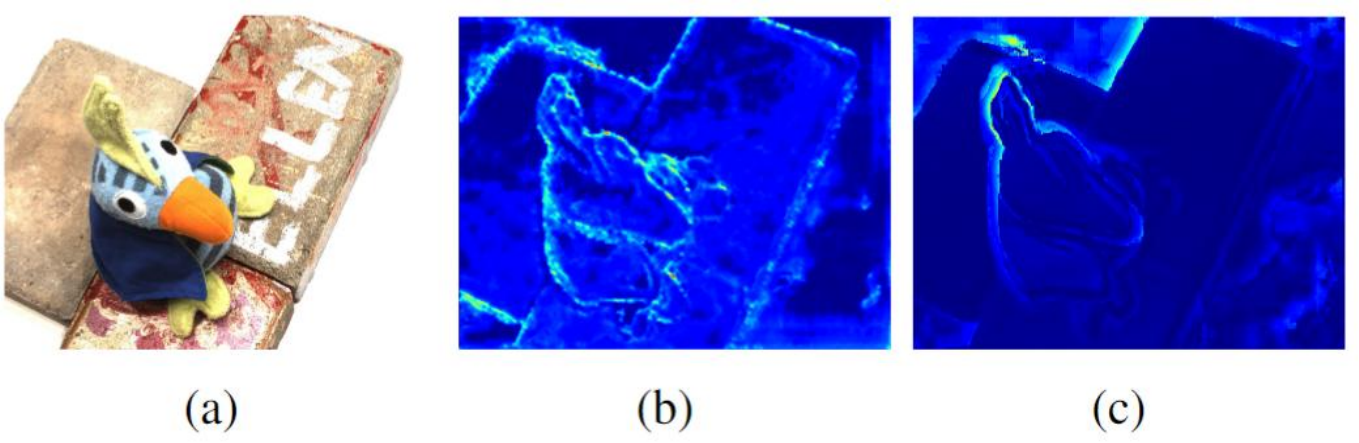}
    \caption{
    (a) Reference image.
    (b) The range map produced by ARNet.
    (c) The eccentricity map, eccentricity between the depth speculated by the GDS module and the previous estimated depth. The brighter the area, the higher the value, and vice versa.}
    \label{range}
\end{figure}

\subsubsection{GRU-based Depth Speculator}
We compare the results on the DTU test set with and without the GDS module. Removing the GDS will reduce the number of depth hypotheses; therefore, to ensure a fair comparison, we compensate for the missing hypotheses at levels 1 and 2 using those estimated in the previous level.
As shown in Table~\ref{tab:gds}, employing GDS-derived depth hypotheses at all levels consistently outperforms the use of compensated hypotheses from preceding levels. This demonstrates that the depth predictions generated by the GDS module more effectively support high-quality depth regression.
\begin{table}[htbp]
    \centering
    \caption{Ablation study concerning the effects of GRU-based Depth Speculator (GDS) module. Symbols \textit{S} and \textit{C} denote configurations with the speculated depth (GDS) and compensated depth, respectively. The number of input views $N$ is set to $5$.}
    \begin{tabular}{cc|ccc}
    \hline
         Level 2 & Level 1 & Acc.(mm)↓ & Comp.(mm)↓ & Overall(mm)↓\\
    \hline
        C & C  & 0.352 & 0.373 & 0.362\\
        S & C  & 0.342 & 0.364 & 0.353\\
        C & S  & 0.336 & 0.363 & 0.349\\
    \hline
        S & S  & \textbf{0.333}& \textbf{0.346}& \textbf{0.340}\\
    \hline
    \end{tabular}
    \label{tab:gds}
\end{table}

\subsubsection{Number of Views}
Finally, we assess the influence of the number of input views $N$ on the performance of ZipMVS. Table \ref{num_views} presents results in the DTU dataset. Increasing the number of views helps resolve occlusions and generally leads to higher-quality reconstructions. However, performance saturates after $N=4$, while additional views incur higher GPU memory consumption and longer inference times. Notably, although our method reaches saturation at $N=4$, most competing approaches require five input views to achieve comparable performance.

\begin{table}[htbp]
    \centering
    \caption{Ablation study on the number of input views N on DTU.}
    \label{num_views}
        \begin{tabular}{c|ccc|cc}
        \hline
             N &  Acc.(mm)↓&  Comp.(mm)↓& Overall(mm)↓ & Mem.(MB) & Runtime(s)\\
        \hline
             2 & 0.4617 &  0.2424& 0.3520& \textbf{1102}& \textbf{0.075}\\
             3 & 0.4650 & \textbf{0.2354} & 0.3502& 1175& 0.088\\
             4 &  0.3692&  0.2840& \textbf{0.3266}& 1250& 0.096\\
             5 &  \textbf{0.3331}&  0.3460& 0.3396& 1322& 0.112\\
             6 &  0.3341&  0.3448& 0.3395& 1396& 0.125\\
        \hline
        \end{tabular}
\end{table}

\section{CONCLUSION}
This work presents ZipMVS, an efficient multi-view stereo framework designed to meet the computational constraints of real-world scenarios such as aerospace applications.
The central contribution of our method is a novel differentiable depth-sampling strategy that allows each pixel to adaptively refine its own search range and propagate depth hypotheses across levels. This pixel-adaptive mechanism leads to a substantial reduction in cost volume size while preserving high reconstruction fidelity.
Experiments on DTU and Tanks and Temples demonstrate that ZipMVS achieves competitive accuracy while using substantially less memory than non-iterative baselines, and offers a favorable trade-off compared to iterative methods.
This highlights the intrinsic strength of the sampling mechanism itself, which avoids the high computational costs while remaining competitive with other efficiency-oriented methods.

\section*{ACKNOWLEDGMENTS}

The first author gratefully acknowledges Prof. Hongshan Yu, Prof. Javier Civera and Prof. Zhaoxin Li for their discussions and manuscript refinements that improved this work. The core of this research was completed before November 2025.

\end{document}